\documentclass[10pt,journal,compsoc]{IEEEtran}
\usepackage{amssymb}
\usepackage[leqno]{amsmath}
\usepackage[pdftex]{graphicx}
\usepackage{epsfig}
\usepackage{epstopdf}
\usepackage{algorithm}
\usepackage{algorithmic}
\usepackage{array}
\usepackage{caption,subcaption}
\usepackage{color}
\usepackage{float}
\usepackage{multirow}
\usepackage{footmisc}
\usepackage{picinpar}
\usepackage{longtable,pdflscape}
\usepackage{amsmath}
\usepackage{amssymb}
\usepackage{ragged2e}
\RequirePackage{xcolor}[2.11]
\colorlet{siaminlinkcolor}{green!50!black}
\colorlet{siamexlinkcolor}{red!60!black}
\colorlet{siamreviewcolor}{black!50}

\usepackage[colorlinks=true,
    linkcolor=cyan,
    citecolor=siamexlinkcolor,      
    urlcolor=cyan]
    {hyperref}
 \usepackage{footmisc}
\usepackage{fancyhdr}
\usepackage{rotating}
\usepackage[noabbrev,capitalise,nameinlink]{cleveref}
\graphicspath{ {./images/} }
\makeatletter
\@addtoreset{footnote}{section}
\usepackage{indentfirst}
\ifCLASSOPTIONcompsoc
  \usepackage[nocompress]{cite}
\else
  \usepackage{cite}
\fi
\ifCLASSINFOpdf
\else
\fi
\begin{document}
%
\title{Accretionary Learning with Deep Neural Networks}
%
%
%

\author{Xinyu Wei,
        Biing-Hwang Fred Juang,
        Ouya Wang,
        Shenglong Zhou
        and Geoffrey Ye Li
\IEEEcompsocitemizethanks{\IEEEcompsocthanksitem X. Wei and BHF. Juang are with the School
of Electrical and Computer Engineering, Georgia Institute of Technology, Atlanta,
GA, 30332.\protect\\
Email: juang@ece.gatech.edu.
\IEEEcompsocthanksitem O. Wang ,S. Zhou and GY. Li are with the Department of Electrical and Electronic Engineering, Imperial College London, London, SW7 2BX.\protect\\
Email: ouya.wang20, shenglong.zhou, geoffrey.li\}@imperial.ac.uk.}
}

%
%

\markboth{~}
{Shell \MakeLowercase{\textit{et al.}}: Bare Demo of IEEEtran.cls for Computer Society Journals}
%



\IEEEtitleabstractindextext{\justify
\begin{abstract}
One of the fundamental limitations of Deep Neural Networks (DNN) is its inability to acquire and accumulate new cognitive capabilities. When some new data appears, such as new object classes that are not in the prescribed set of objects being recognized, a conventional DNN would not be able to recognize them due to the fundamental formulation that it takes. The current solution is typically to re-design and re-learn the entire network, perhaps with a new configuration, from a newly expanded dataset to accommodate new knowledge. This process is quite different from that of a human learner. In this paper, we propose a new learning method named Accretionary Learning (AL) to emulate human learning, in that the set of objects to be recognized may not be pre-specified. The corresponding learning structure is modularized, which can dynamically expand to register and use new knowledge. During accretionary learning, the learning process does not require the system to be totally re-designed and re-trained as the set of objects grows in size. The proposed DNN structure does not forget previous knowledge when learning to recognize new data classes. We show that the new structure and the design methodology lead to a system that can grow to cope with increased cognitive complexity while providing stable and superior overall performance.
\end{abstract}

}

\maketitle

\IEEEdisplaynontitleabstractindextext

%
\IEEEpeerreviewmaketitle

\IEEEraisesectionheading{\section{Introduction}\label{sec:introduction}}

%
%
%
%
\IEEEPARstart{O}{ne} of the current focuses in machine intelligence research is pattern recognition. A machine that can recognize an object or a pattern is considered to be performing a certain intelligent, i.e., human-like capability. Pattern recognition as a technical problem nonetheless has been around for much longer than the notion of artificial intelligence. Many earlier formulations of the pattern recognition problem exist, the most systematic of which is perhaps Thomas Bayes' optimal decision theory \cite{duda2001pattern}, which further led to the area of statistical pattern recognition \cite{svensen2007pattern}. Systems that attempt to realize statistical pattern recognition in real applications are abundant. 
 
In the past decade, Deep Neural Networks (DNN), with new ideas from machine learning and statistical estimation, have significantly enhanced the recognition performance and started to enjoy broad applications in areas such as image classification, speech recognition, natural language processing \cite{schmidhuber2012multi, deng2013new}, and wireless communications recently \cite{qin2019deep, ye2020deep}. Applications of DNN in other areas are also plenty. 

Differences aside, DNNs do share with the conventional statistical pattern recognition approach a common setup of the problem; to wit, an unknown observed pattern is to be recognized as one of the $M$ known classes. We shall call this collection of recognizable objects or patterns the recognition set, which has a size of $M$. This may be termed as a closed-set recognition task, in which the recognition set and its size are fixed. A system designed or trained to recognize $M$ classes cannot be readily used to recognize any additional classes. (Some non-rigorous practices that heuristically deviate from the original formulation do exist. For example, the idea of out-of-vocabulary rejection is prevalent in speech recognition to handle speech input that contains words not in the original vocabulary when the recognizer was designed. These supplementary measures cannot be systematically analyzed with the original formulation of Bayes and are thus excluded for consideration here.) In contrast, human cognitive capabilities are learned without the rigid ``closed-set" constraints.  

Instead, humans learn their cognitive capabilities mostly in an ``open-set" fashion in which new classes of patterns emerge and are ready to be learned without having to be absorbed into a prescribed set. The learner learns the additional cognitive capability mostly cumulatively. As an illustrative example, let us consider how a child is being taught to recognize Arabic numerals. Most likely, the teacher (or parent) will start with a digit, say 1, and show examples for the child to identify. Afterwards, the child continues to learn other digits. The cognitive capability is  learned in a cumulative manner. To the best of our knowledge, no child is being told that he or she will be learning 10 digits prescriptively before starting to learn to identify any digit. This is obviously different from the fundamental formulation of Bayes' decision theory, which starts with a well defined recognition set, a pre-requisite that cannot be avoided. 

Researchers have been working on explaining human learning behavior for many years. A notable theory by Norman \cite{norman1978notes} and Rumelhart \cite{rumelhart1976accretion} postulates that humans have three learning modes, namely, accretion, tuning, and restructuring. Accretion is the process of adding new knowledge to the current knowledge frame of an open-set task. For example, learning to recognize letters in an alphabet involves a process of knowledge accretion. During learning, people do not learn all letters in the alphabet at once, but do rather encounter new letters incrementally. Information necessary for the identification of new letters is added to the current knowledge frame; the sense of “size” of the alphabet is rarely a crucial prescribed part of the accretion process. Tuning is the process of adjusting some variables in the current knowledge frame to improve the generalizability and the performance of the learned capability. Restructuring is to build a new knowledge frame for storing the newly acquired information conditioned on the current knowledge frame. These three modes do not work independently but co-occur during the learning process, which ensures the continual and stable learning performance of human beings. The three modes support each other to form the overall learning process, which, for simplicity, shall be called Accretionary Learning in this paper. Here, we focus on realizing accretion learning with computational models, a primary candidate of which is the artificial neural networks. 

According to Norman \cite{rumelhart1976accretion} and Rumelhart \cite{norman1978notes}, accretion learning has the following properties: 1) An existing knowledge framework for analyzing data and storing knowledge of new data classes is required; 2) The number of data categories that the model recognizes is not fixed and can increase as new data classes arise; 3) Learning new data categories will not cause an obvious degradation or interference of the previous knowledge. For a corresponding computational model that realizes these properties and achieves the goal of accretionary learning, it should have the following features: 1) The model contains a base structure for analyzing input data and categorizing pattern classes; 2) The model has a dynamic structure that can expand itself for storing new information to cope with a growing recognition set; 3) The previously learned structure (i.e., internal parameters and construct) should not be grossly affected during the accretion of new knowledge, to maintain stability in learning and the associated result. We choose DNNs as the computational structure to realize this learning model because they have been frequently used as feature extractors and classifiers. In addition, as we will show, this accretionary learning model also works well in closed-set tasks with fixed recognition sets, comparable to that of the conventional DNNs. 

As a key element in realizing accretionary learning, we propose a detection-based learning scheme following the feature mentioned above. The working mechanism of the model includes a process of individual pattern detection and overall decision-making. The process consists of two modular systems: 1) an individual class detection system, and 2) a decision rendering system based on class discrimination. When an unknown pattern is observed as input to the system, the detection system uses the current knowledge to determine the individual degree of association with each and every of those learned pattern classes. Then the decision rendering system abstracts the intermediate feature formed by the array of degrees of association into a class label as the decision. During accretionary learning, the two systems are trained separately to maintain growth flexibility but can be further jointly optimized, if necessary, to emulate the aforementioned tuning and restructuring modes. When the recognition set grows together with the supplied training data, the detection system of the model first expands itself by forming a new detector of the unseen pattern class, and the decision rendering system then modifies its structure to accommodate the new knowledge. The two systems together ensure the realization of stable and efficient accretionary learning. 

This paper is organized as follows. In the next section, we review the knowledge of incremental learning and three categories of learning strategies for DNNs. The design philosophy and structure of the accretionary learning model are introduced in Section \ref{sec:DESIGN}. In Section \ref{sec:Expriment}, we produce the testing accuracy of the accretionary learning model with different bootstrap sizes and learning orders, together with the performance comparison with the traditional CNN. We conclude the paper in Section \ref{sec:Conclusion}  

 

\section{RELATED WORKS} \label{sec:RELATED WORKS}
Researchers have made attempts at making DNNs learn to deal with the “open-set” recognition problem. The term “incremental learning” is frequently used to describe the paradigm of adapting the system to new data or new capabilities \cite{bruzzone1999incremental, diehl2003svm}.  Consider a DNN that has been optimized for a task that involves $M$ classes of patterns. When the recognition set grows, say from size $M$ to $M+1$, an additional output node representing the new class is added to the neural networks together with all necessary synaptic connections. At this point, the system designer can choose to use the new pattern data to retrain the entire system, or only to train the newly added synaptic connections, leaving the original network structure and the learned parameters largely intact. One main challenge in the first option is called catastrophic forgetting, which was first proposed in \cite{mccloskey1989catastrophic} and refers to the phenomenon that the system forgets the previous knowledge while learning new information. This phenomenon was further analyzed and explained in \cite{french1999catastrophic, kemker2018measuring}. According to the experiments in these works, the system would experience an obvious and serious degradation in recognizing the previous pattern classes. In order to eliminate or alleviate this phenomenon, researchers have designed different structural DNNs with various learning strategies, which can be divided into three main categories: regularization, rehearsal, and dynamic structure. 

\subsection*{\it A. Regularization} 
The regularization strategy \cite{li2017learning, kirkpatrick2017overcoming, zenke2017continual, dhar2019learning} overcomes catastrophic forgetting by adding a regularization term to the cost function of an incremental learning system, which prevents the weights in the learning system from being updated towards the direction that only optimizes the performance on the new classes. LwF \cite{li2017learning} is one of the representative algorithms and applies knowledge distillation \cite{hinton2015distilling} to consolidate the previous memory while learning to recognize new pattern classes. Besides LwF, other algorithms like EwC \cite{kirkpatrick2017overcoming} and SI \cite{zenke2017continual} prevent the old memory from being overwritten by first finding the weights that are important to old knowledge and then adding regularization terms to protect them from being changed drastically during the process of learning new pattern classes. 

\subsection*{\it B. Rehearsal} 
The Rehearsal strategy uses old data to consolidate the previous memory. When learning new pattern classes, system models are trained on the datasets that contain both new and old data. Depending on whether to use the real old data, the rehearsal methods can be divided into two types: regular rehearsal and pseudo rehearsal. 

Regular rehearsal uses part of the real old data to help the learning model to avoid catastrophic forgetting. iCarL \cite{rebuffi2017icarl} is one representative regular rehearsal algorithm and achieves considerable performance. The learning system of iCarL contains a memory buffer for storing the old data. When a new data class arises, the old data in the memory buffer and the new data will form a new training set for the system to learn the new classes as well as to consolidate the previous knowledge. After being well trained, the learning system will dynamically modify the composition of its memory buffer by removing less important old data and adding important new data.

Unlike regular rehearsal, the pseudo rehearsal generates fake old data according to the distributions of old data classes and does not need extra memory for storing previous data. This idea was first proposed in \cite{robins1995catastrophic} and then realized with different methods \cite{draelos2017neurogenesis, kemker2017fearnet, rios2018closed, wu2018incremental}. \cite{draelos2017neurogenesis, kemker2017fearnet} use autoencoders to reproduce the old data and to model the distribution of their latent features. Besides autoencoder, the Generative Adversarial Networks \cite{rios2018closed, wu2018incremental} is another tool for generating old data. 

\subsection*{\it C. Dynamic Structure} 
A dynamic structural learning model expands itself with extra neural networks to accommodate the knowledge of the new data classes, and only the weights of the expanded neural networks get updated when the model is learning new data classes. This strategy not only allows the model to acquire knowledge of the new data classes but also avoids interference on the previous memory. \cite{sarwar2019incremental} designs a partially shared network that can continuously learn new data classes. This shared network consists of two components: a shared network and branch networks. The shared network is used to extract common features from the input data, and the branch networks are used for recognizing the class. When new data classes arise, a corresponding new branch network will be linked to the shared network and is trained to recognize new data classes. During training, only the weights in the branch network get updated. This learning structure performs well in ImageNet. 

Although the methods mentioned above provide ideas for building a continual and stable learning model, the design concept is ad hoc and lacks a general human-like learning manner to produce consistent learning results. In order to build a human-like learning model, we thus adopt the accretionary learning theory of Norman and Rumelhart and propose below a design paradigm as well as a computational learning process to be realized by artificial neural networks. 
\section{ACCRETIONARY SYSTEM DESIGN}\label{sec:DESIGN}
Unlike traditional machine learning algorithms that train models for closed-set tasks with prescribed recognition sets, humans learn in an open environment where the number of pattern classes is not necessarily predefined, and the size of the dataset for human learning is often floating. In real-world scenarios, humans interact with various information every day and absorb new knowledge from the presented information continually and accumulatively. To emulate this human-like learning behavior, we follow the learning theory of Norman and Rumelhart and propose a system design methodology to realize accretionary learning. Our proposed learning paradigm integrates the coordination and interworking of the three learning modes of Norman and Rumelhart as a whole. In this section, we analyze and discuss the key design methodology that has the potential of realizing accretionary learning based on artificial neural networks.
\subsection{Design Philosophy}
\begin{figure}[!h]
\centering
\includegraphics[width=0.5\textwidth]{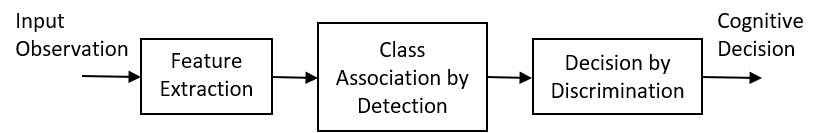}

\caption{Conceptual blocks in accretionary learning}
\label{fig:part3-1}
\end{figure}

The design philosophy below describes how the accretionary learning model interacts with external information like images or sounds and gain new knowledge accumulatively. The proposed working mechanism of the learning model involves three steps, which are feature extraction, association by detection, and discrimination-based cognitive decision, as depicted in Fig. \ref{fig:part3-1}.

\subsubsection{Feature Extraction}
A human uses sensory receptors to receive external information and to transfer the information into various neural signals. For example, an image will activate the receptors in the eyes and then be transformed into neural signals for the brain, including the visual cortex, to analyze to eventually form visual perception. To draw a parallel, the accretionary learning model is also equipped with a module for preprocessing the input data like images or sounds into features. In terms of cognitive functions, the feature extraction stage may be considered \textit{object independent} and does not have to involve \textit{the abstract notion of object class identity that is logically or cognitively assigned according to the given task}. The feature extraction stage is not particularly different from many traditional designs, including those that use spectral representations as features, or modern ones like restrictive Boltzmann machines or autoencoders that produce a representation of certain universal statistical regularity (i.e., non-class-dependent, under a general homogeneity assumption on the data) as the feature. In this paper, the module is realized by Convolutional Neural Networks (CNN), which are useful tools for capturing features from the visual input. Other popular neural networks can also be utilized for feature extraction, as in other applications. While we use imagery classes, namely the handwritten digits, as the recognition set in our experiments, the above feature extraction principle applies to other patterns without loss of generality.

\subsubsection{Association by Detection}
After the first stage, the extracted feature from the input sensory data is sent to the brain for analysis and knowledge accretion. At this point of learning, object classes are learned individually. The goal of (human) learning is to capture how the presented feature is critically associated with the particular class that is being learned. In terms of computational model, this can be formulated in a way similar to the statistical detection theory, in that the two hypotheses, i.e., whether the presented feature indicates the presence of the target object class or not, are tested against each other to produce a decision of detection. \textit{This is association by detection as its focus is the target class rather than the "non-target" classes, which may vary dynamically throughout the learning process.} In other words, at this point, the notion of identifying a class is one of an open-set, without making assumptions on what the eventual recognition set will be. The association by detection is learned for each and every data class individually, forming a bank of detectors. This is a key design philosophy of accretionary learning. 

A brief review of hypothesis testing is in order. Let the hypothesis of  presence  of the target object be denoted by $H_0$ and its alternative hypothesis $H_1$. Statistical learning is to make use of the known data to obtain values of the parameter set, $\theta$, that define $p(X|H_0,\theta)$ and $p(X|H_1,\theta)$ such that the likelihood ratio test eventually achieves the optimal performance in decision. The likelihood ratio test is defined as:
\begin{equation*}
     \frac{p(X|H_0,\theta)}{p(X|H_1,\theta)}\lessgtr \tau,
\end{equation*}
where $\tau$ is a prescribed threshold. If $p(X|H_0,\theta)/p(X|H_1,\theta)$ is above the threshold, the $H_0$ hypothesis is accepted, leading to the declaration of presence of the target object in the observed feature. If not, $H_1$ prevails, which means the target object is considered absent in the observation. Note that, the regime of hypothesis testing will produce two types of errors, Type 1, which is often called ``miss", and Type 2, which is called ``false alarm", in radar terminology. The threshold $\tau$ is prescribed based on the consideration of the tradeoff between type 1 and type 2 errors, often represented as an operating point in the Receiver Operating Characteristic (ROC) curve \cite{fawcett2006introduction}.


\begin{figure}[!h]
\centering
\includegraphics[width=0.3\textwidth]{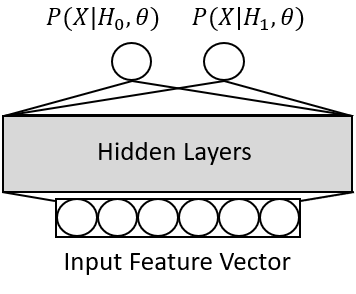}

\caption{The general structure of a detector network for a target object class; It evaluates the likelihood of the presence of the target object and the lack of it in the observation input. When the system learns to deal with multiple classes, a corresponding number of the detectors are individually learned and then operate in parallel}
\label{fig:part3-1-2}
\end{figure}

The regime of hypothesis testing can be realized by a neural network, which produces as output the two likelihood values, $p(X|H_0,\theta)$ and $p(X|H_1,\theta)$, as depicted in Fig.~\ref{fig:part3-1-2}. (For the current exposition, the network is assumed to produce the two likelihood values as output. It can nonetheless be expanded to a more sophisticated testing regime with more than two numbers as output.) The detector network, for a particular target class, is trained to minimize a combination of the two types of errors in a number of ways. Actual choices of the optimization objective will be discussed in the next section. Each detector network is trained to register the representational characteristics particular to the data of the target class \textit{in contrast to the non-target class data} at the time of the detector learning. All the detector networks are separate from each other, and together they form a composite bank of detectors that can analyze the similarity between input data and the already learned object classes via hypothesis testing on the data. 

This composite detector bank will expand one at a time when  a new data class is introduced. During the training of the new detector network, data for the target class representing the null hypothesis as well as selected data for the alternative hypothesis must be presented. We will discuss how the data for an alternative hypothesis is to be provisioned in the next section. Note that when new detectors are learned, parameters for those previously trained detectors remain intact, and thus the corresponding knowledge about previous classes is preserved. The output of the composite detector bank can be viewed as a map of the transformed neural representation for cognitive discrimination in the later stage of the cognitive function. 

\subsubsection{Discrimination-based Decision Network}
\begin{figure}[!h]
\centering
\includegraphics[width=0.3\textwidth]{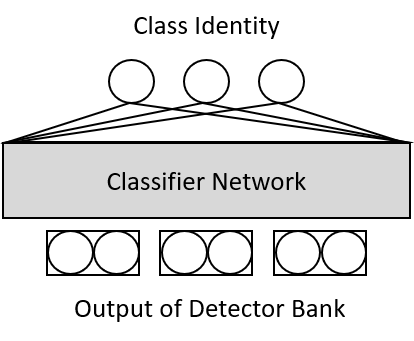}

\caption{The general structure of a classifier decision network, which abstracts the intermediate feature into one of the learned classes. The nodes in the input layer are twice as many as those in the output layer}
\label{fig:part3-1-3}
\end{figure}

While the bank of detectors performs class association, it is not yet trained to discriminate among classes that at times may show a high degree of confusability. After all, each detector is trained separately. To render a sensible recognition decision, a classifier capable of discerning the differences in the entire computed set of likelihoods for all the learned classes is both necessary and helpful. Take handwritten digits as an example. Digit ‘7’ shares a substantial similarity with digit ‘1’. The outputs of the detector for ‘1’ and ‘7’ may show potentially confusable similarity. However, since ‘7’ may also show higher similarity to digit ‘9’ than ‘1’ would to ‘9’, the whole array of outputs of the detector bank thus presents additional evidence for discernment between ‘1’ and ‘7’. This discriminating classifier can be accomplished by another neural network, as is conventionally done in supervised multi-class learning. Fig.~\ref{fig:part3-1-3} depicts a basic structure of the network whose input is the results of the detection stage and the output layer has the same number of nodes as the number of learned classes. When a new object class is introduced and the corresponding new detector is trained, the discrimination network is incremented by two new nodes at the input layer, to accept the results of the new detector, and by one new node at the output, to represent the new class. The process mimics the tuning mode of accretionary learning.

\subsection{Realization of the Accretionary Learning Paradigm}
\begin{figure}[h!]
\centering
\includegraphics[width=0.5\textwidth]{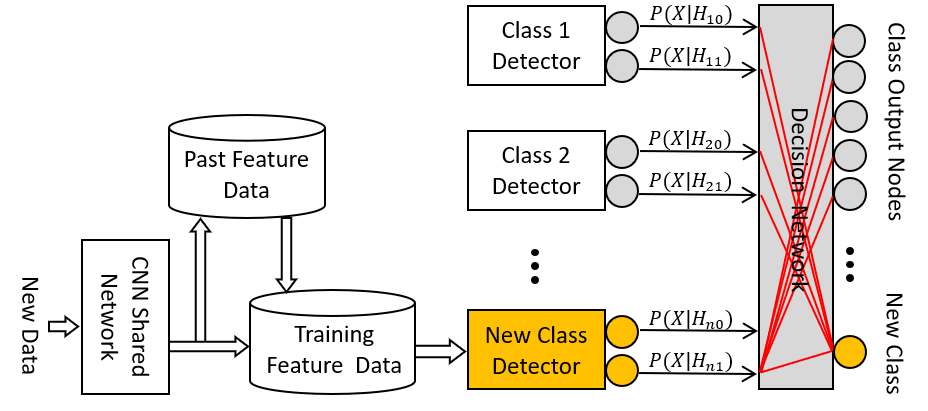}
\caption{The overall structure of accretionary learning, integrating aforementioned modules. The newly added components, the detector and the associated connections marked in color, are to store new knowledge when learning a new object class.}
\label{fig:part3-2}
\end{figure}

Following the design philosophy above, we propose a detection-based learning model that emulates accretionary learning. Fig.~\ref{fig:part3-2} shows the structure of the model and depicts the process of accretionary learning. The shared network, which serves as {a frontend} to extract feature data from the input observation, is obvious and omitted in the figure, although an explanation is included in the following section. The data pool represents a store of randomly sampled past tokens. Note that the data pool can be considered as a collection of empirical data, albeit limited, as human memory would often function $p(X|H_0)$.

\subsubsection{Shared Network for Feature Extraction}
Like human's sensory receptors such as eyes and ears, which interact with {real-world information}, the shared network is responsible for preprocessing the sensory input data and transferring them to standard-format features for further analysis. In this study, as alluded to earlier, we opt for the Convolutional Neural Networks (CNN) as the preprocessing, feature-extracting shared network. The shared network contains multiple convolutional layers, which are effective at producing feature representations from the given images. The feature output of the shared network is class-independent. To build an  {effectively shared} network, we need to pay attention to the structure of the network such as its number of layers, the size of the kernel for each convolutional layer, and the dimensionality of the extracted feature. We will report the exact configuration of the shared network in the experiment section. 

\subsection*{\it  Mean-Squared Error} 
\begin{itemize}
\item The Mean-Squared Error (MSE) measures the average of the squares of errors. In the current context, the detector can be viewed as a 2-class (binary) recognizer with two output nodes, one representing the target class and the other representing "the rest of world" (ROW) other than the target. The objective is often used as the loss function in the supervised training of a conventional neural networks recognizer with the error back-propagation (EBP) algorithm. In such a binary classification problem, typically, when a learning token of the target class is presented, the target and the ROW output values are expected to attain the maximum value of 1 and the minimum value of 0, respectively; when a learning token of ROW is presented, the expected output would be reversed to 0 and 1, respectively. Let $x_i$, $y_i$ be the expected output vector and the computed output vector, respectively, evaluated on the $i^{th}$ token. The loss function is defined as 
\begin{equation}
    L_{MSE}=\frac{1}{n}\sum_i (x_i-y_i)^2,
\end{equation}
where $n$ is the batch size of the training tokens. The network parameters are optimized by minimizing the above loss function, usually through the EBP algorithm.
\end{itemize}

\subsection*{\it  Cross-Entropy}
\begin{itemize}
\item The cross-entropy (CE) measures the relative entropy between two probability distributions, numerical evaluation of which is typically performed over an identical set of sampled events. In a binary classification problem, the two output values of a deep learning model, such as a neural network, are viewed as the likelihoods that the input pattern is from the respective classes, i.e., the target and the ROW. When a training token of the target class is presented, the expected pair of likelihood values are set as $y$ and $1-y$, where $y$ is often chosen to be 1 or very close to 1. With $\hat{y}$ representing the computed value of the network for the target likelihood, the cross-entropy loss function is given as:
\begin{equation}
    L_{CE}=\sum_i \Big(y\log(\hat{y})+(1-y)\log(1-\hat{y})\Big),
\end{equation}
where {$y$ and $1-y$} are the expected output for the given data, and {$\hat{y}$ and $1-\hat{y}$} are the computed output of the network evaluated on the given data. The network parameters are optimized, e.g., with the gradient descent algorithm, to iteratively minimize the above loss function. Similar to the MSE case,  {the summation is over a batch} of labeled learning tokens.
\end{itemize}

\subsection*{\it  Area Under the ROC Curve (AUC)}
\begin{itemize}
\item Since every detector network conducts hypothesis testing on the input data, an ROC curve can depict the relationship between a model’s True Positive Rate (TPR) and False Positive Rate (FPR) at various thresholds or operating points. The AUC, referring to the area under an ROC curve, reflects the model’s performance. A larger AUC of a model means a better performance it has. One popular way to calculate AUC is the Mann-Whitney statistic, which is denoted as: 
\begin{eqnarray}
    A=\frac{\sum_{i=1}^{m} \sum_{j=1}^{n} 1_{x_i\textgreater y_i}}{mn},
\end{eqnarray}
where $x_1,\ldots,x_m$ are the output of a classifier on the positive samples, and $y_1,\ldots,y_m$ are the output of the classifier on the negative samples. In order to improve the model’s performance, we need to maximize the AUC.
\end{itemize}

There can be other choices of learning objectives for the optimization of the detector networks. In our experiments, we use the cross-entropy to compute the loss of every output of each detector network. Before training the detector network for a particular pattern class, we label positive samples as (1,0) and the negative samples as (0,1). The two computed output values, representing the results of $H_0$ and $H_1$ test, are denoted as $p(X|H_0,\theta)$ and $p(X|H_1,\theta)$, which take the roles of $\hat{y_s}$. We also denote the $i^{th}$ data sample as $x_i$, the true likelihood of the sample data as {$y_0$ and $y_1$}, and the total loss as:  
\begin{eqnarray*}
\arraycolsep=0pt\def\arraystretch{1.75}
\begin{array}{rll}
Loss=-\sum_{i=1}^m \Big[&\hat{y_0}\log(p(x_i|H_0,\theta))+ \\
    &(1-\hat{y_0})\log(1-p(x_i|H_0,\theta))+\nonumber\\ 
    &\hat{y_1}\log(p(x_i|H_1,\theta))+\\
    &(1-\hat{y_1})\log(1-p(x_i|H_1,\theta))\Big].
\end{array}  
\end{eqnarray*}

When a new pattern class appears, we create a new detector network for analyzing and determining the degree of association between a given pattern and the new class. The new data and the data from the data pool form the training set for this detector network. The new detector network has the same training strategy as the previous detector networks. During the training process, the parameters of the existing networks are kept intact, which protects the learned knowledge from being unduly affected. The use and management of the data pool will be explained below in Section \ref{sec:data-pool}.
\subsubsection{Discriminatory Decision Network}
The outputs of all the detector networks form an intermediate layer of cognitive feature representing the model’s overall impression on the input sensory data before the final cognitive decision is rendered. The discriminative network will abstract the intermediate feature into one of the learned class labels. The decision network shown in Fig.~\ref{fig:part3-2} is a single-layer MLP (during this initial study), and we use cross-entropy as its loss function for performance optimization. When a new pattern class is introduced, the discriminated network is expanded by adding two input nodes to accommodate the new outputs from the new detector network and adding an output node representing the new class. During training, only the weights of new connections are updated. 
\subsubsection{Sampled Empirical Data Pool}\label{sec:data-pool}
Like humans that have memory of previously seen objects, the accretionary learning model also contains a data pool for storing some already learned data. Data in the data pool is randomly sampled and used for training new detector networks and the expanded discriminated network when new pattern classes arise for learning. After completing each epoch of accretionary learning, data of the newly emerged classes will be stored in the data pool.  

One immediate issue of interest is how much diversity of data that need to be presented during the learning of the initial pattern classes and their detectors. We call this the bootstrap size, in terms of the number of initial classes, the data of which will be involved in the optimization of detectors. Since all learned networks will remain intact during accretionary learning as the pattern classes grow, there is a concern that a serious lack of diversity in training the first set of detectors may lead to a weak response in dealing with newly emerged pattern classes. We investigate this concern in the experiment section under the topic of bootstrap size. 

Since the data pool has a limited capacity for storing data, we will continue to address effective strategies of preserving data for accretionary learning in future work. 
\section{Experiment}\label{sec:Expriment}
To verify the effectiveness of the accretionary learning paradigm, including the proposed network structure and the learning process, we design assessment experiments around the MNIST dataset. Using human learning as a reference, we are motivated by the parallel of teaching a child to recognize the digits, not 10 digits all at once but one or a few at a time until all digits have been learned. We record the system performance as it accumulates the learned capability sequentially so as to monitor the accretionary behavior of the system. Note that we have in Section \ref{sec:DESIGN} laid out many possible implementational variations (e.g., the detector learning criteria and so on), but in our current studies, no attempt was made to compare these variations. They will be investigated in the future. 

The first question to answer is how well the proposed accretionary learning, which we believe is following the human learning process, is going to perform. This question has two relevant perspectives, one in terms of its behavior in the growing cognitive capability (will it fail to grow before all digits are taught?) and the other in terms of its performance relative to the current state-of-the-art systems, which is while deemed successful in performance but nonetheless lacking the growing capability. The second question is if the learning order affects the learning results. The learning order refers to the order of data/object classes that the system learns with. If the learning process is indeed following the above-reasoned development goal of performing intelligent tasks in an open-set environment, we expect its performance to be mostly independent of the class learning order. If the accretionary learning can only learn new object classes in a specific order, the proposed learning algorithm cannot be considered practical because new classes in real-world learning usually appear randomly. The third question, as alluded to earlier, is how the bootstrap size would impact the overall performance. The bootstrap size refers to the initial knowledge that the accretionary learning system acquires through non-incremental learning. The key consideration is about the variety classes of data that are involved in the initial stage of learning. If the bootstrap size is too small, the model may face difficulty in obtaining new knowledge  {because of the initial exposure to data,  and thus} the learned knowledge may be considered weak. 

\begin{figure}[!th]
\centering
\begin{subfigure}{.495\textwidth}
	\centering
	\includegraphics[width=.5\linewidth]{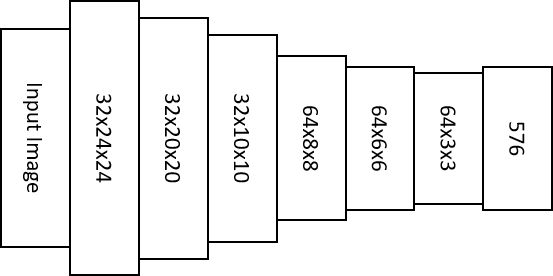}
	\caption{Shared CNN network}
	\label{fig:part4_1}
\end{subfigure}	 
\begin{subfigure}{.495\textwidth}
	\centering
	\includegraphics[width=.5\linewidth]{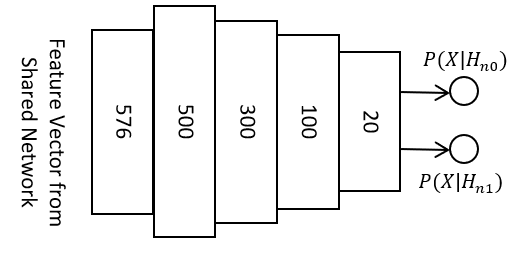}
	\caption{Single class detector network}
	\label{fig:part4_1_2}
\end{subfigure} 
\caption{The structures of shared CNN network and  single class detector network in current study.\label{fig:Learning_order}}
\end{figure}

In terms of the configuration for the network components, Fig.~\ref{fig:part4_1} and Fig. \ref{fig:part4_1_2} depict the shared CNN network, for feature transformation and extraction and the single-class detector network, respectively.   {The single-class detector network in the following two experiments has the same structure}. The decision network is a two-layer perceptron that abstracts intermediate features from the detector bank into one of the learned classes. 

%
\subsection{System Performance and Bootstrap Size}
Our first experiment is to investigate the overall system performance in terms of the recognition accuracy. For this experiment, we follow the natural numerical order from 0 to 9 as the learning sequence. We design the system with an increasing bootstrap size, starting with a size of 3 and ending at 9. That is, the first system is trained on digits 0, 1, and 2 and then followed by the accretionary learning process thereafter until it reaches the final digit 9. In other words, we start with an initial bootstrapped system using first $M$ classes of data and then perform accretionary learning for the remaining digits, one at a time to  {$M+1, M+2$,} and so on till the final digit. The amount of training data per class is 5000 and a separate evaluation set of size 800 independent of the training set is used to obtain the prediction accuracy, which is an average over all involved classes. 

\begin{figure}[h!]
\centering
\includegraphics[width=0.4\textwidth]{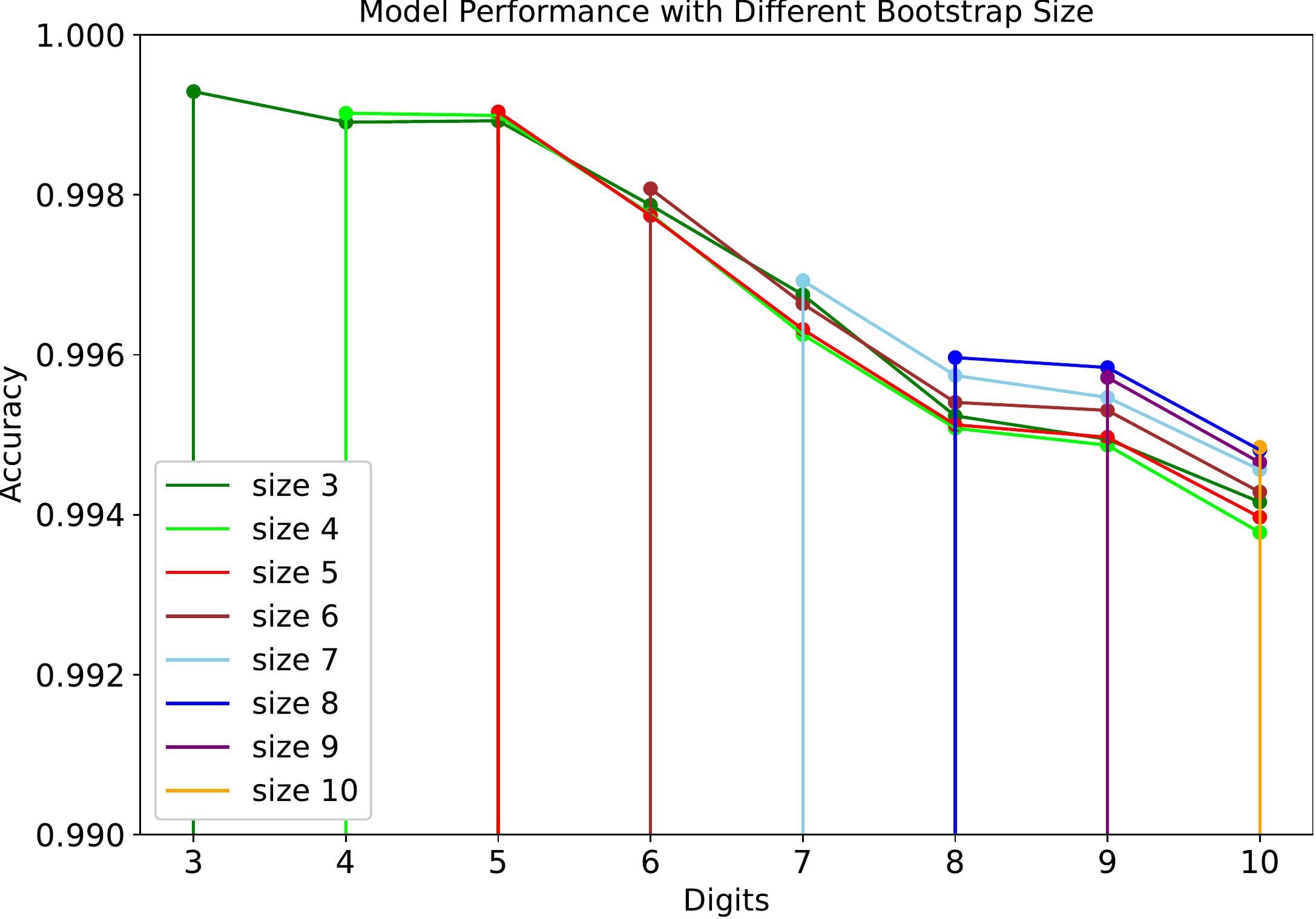}
\caption{The performance of accretionary learning models with different bootstrap sizes.}
\label{fig:Bootstrap}
\end{figure}

The results are plotted in Fig.\ref{fig:Bootstrap}. For example, the initial system with bootstrap size 3, where the decision network is fully trained, shows an accuracy of 0.9993. The fourth digit is then learned according to the accretionary learning process, resulting in an average accuracy 0.9987. The addition of the fifth digit does not seem to change the average accuracy. Afterwards, as the recognition set expands and complexity increases, the recognition accuracy appears to trend downward although only slightly, mostly still above 0.994. This performance is remarkably competitive to the state-of-the-art systems \cite{simonyan2014very}, which can only be designed and optimized for the entire set of ten digits. The figure includes the results for bootstrap size 4 and beyond as well. The overall results are rather consistent. 
\subsection{Learning Order}
To verify that the system performance does not depend on the learning order, we randomly permute the digit sequence as the learning order in lieu of the natural order. To properly manage this experiment, we fix the bootstrap size to five, leaving the other five for accretionary learning. The choice and the order of the two 5-digit sets are randomly sampled. Table \ref{table:1} shows examples of the permutation results. 
\begin{table*}[!th]
\renewcommand{\arraystretch}{1.5}\addtolength{\tabcolsep}{4pt}
	\begin{center}
		\caption{Ten different learning orders for ten bootstrap sets}
    \label{table:1}
		\begin{tabular}{l|c|c|c|c|c|c|c|c|c } \hline 
        Order & 0,1,2,3,4  & 1,9,3,0,7 & 2,8,7,4,1  & 4,9,6,8,1 & 5,0,4,3,7 & 6,2,0,8,4 & 7,3,1,5,8 & 8,7,2,1,4 & 9,3,4,5,6 \\
        \hline

       order1 & 9,5,8,7,6 &2,5,6,8,4&9,5,0,3,6&5,2,0,3,7  &2,6,9,8,1 &1,3,9,5,7&6,0,2,4,9&9,3,5,6,0&0,2,1,7,8  \\ 
       order2 & 8,6,7,9,5&5,8,2,4,6&5,9,3,6,0&2,0,5,7,3 &6,9,2,1,8&3,1,9,7,5&0,6,2,9,4&3,5,9,0,6&2,7,1,8,0 \\
       order3 &7,5,8,6,9&6,5,8,4,2&0,5,3,6,9&0,2,3,7,5 &9,2,8,1,6&1,3,5,7,9&2,6,4,9,0&5,9,6,0,3&1,7,0,2,8 \\
       order4 &5,9,7,8,6&8,5,4,2,6&3,9,6,5,0&3,0,7,2,5  &8,6,9,1,2&5,1,7,9,3&4,9,2,0,6&6,5,0,3,9&7,0,8,2,1  \\
       order5 &8,9,5,6,7&4,6,8,2,5&6,0,9,5,3&7,3,5,2,0 &1,9,6,2,8&7,1,3,9,5&9,6,2,0,4&0,6,3,9,5&8,1,7,0,2 \\
       order6 &7,6,9,8,5&5,2,6,4,8&5,3,0,6,9&2,3,5,7,0 &6,8,9,1,2&3,7,5,9,1&0,4,6,9,2&3,9,6,5,0&0,7,1,8,2 \\
       order7 &7,5,8,6,9&6,2,8,4,5&0,9,3,6,5&0,2,3,5,7 &2,9,8,1,6&1,9,5,3,7&2,0,4,6,9&5,6,9,3,0&2,0,8,1,7 \\
       order8 &7,9,6,5,8&8,6,4,2,5&3,5,6,9,0&3,7,2,0,5 &8,1,6,2,9&5,1,3,7,9&4,9,0,2,6&6,9,0,5,3&7,1,0,2,8 \\
       order9 &9,7,6,8,5&4,5,2,8,6&6,0,5,9,3&7,3,5,2,0  &1,8,6,9,2&7,9,1,5,3&9,6,4,2,0&0,9,3,5,6&8,0,2,7,1  \\  
       order10 &5,6,7,8,9&4,8,6,5,2&6,3,9,5,0&7,0,2,5,3  &1,2,6,8,9&7,5,9,3,1&9,4,0,2,6&0,3,5,9,6&0,1,2,8,7  \\
       
       \hline
 		\end{tabular}
	\end{center}

\end{table*}    
%
%
%

\begin{figure*}[!th]
\centering
\begin{subfigure}{.32\textwidth}
	\centering
	\includegraphics[width=.9\linewidth]{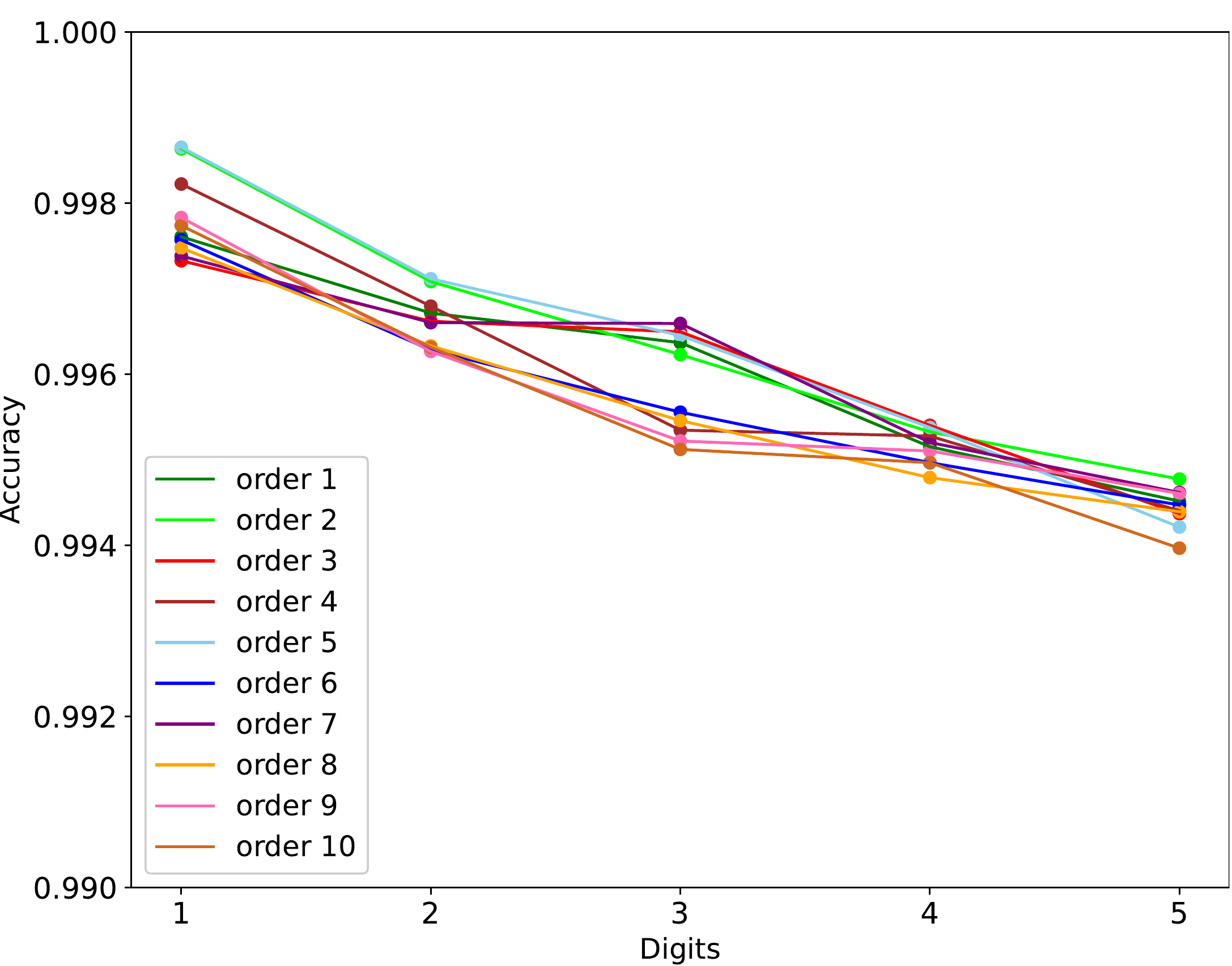} 
	\caption{Bootstrap set $[0,1,2,3,4]$}
	\label{fig:01234}
\end{subfigure}	 
\begin{subfigure}{.32\textwidth}
	\centering
	\includegraphics[width=.9\linewidth]{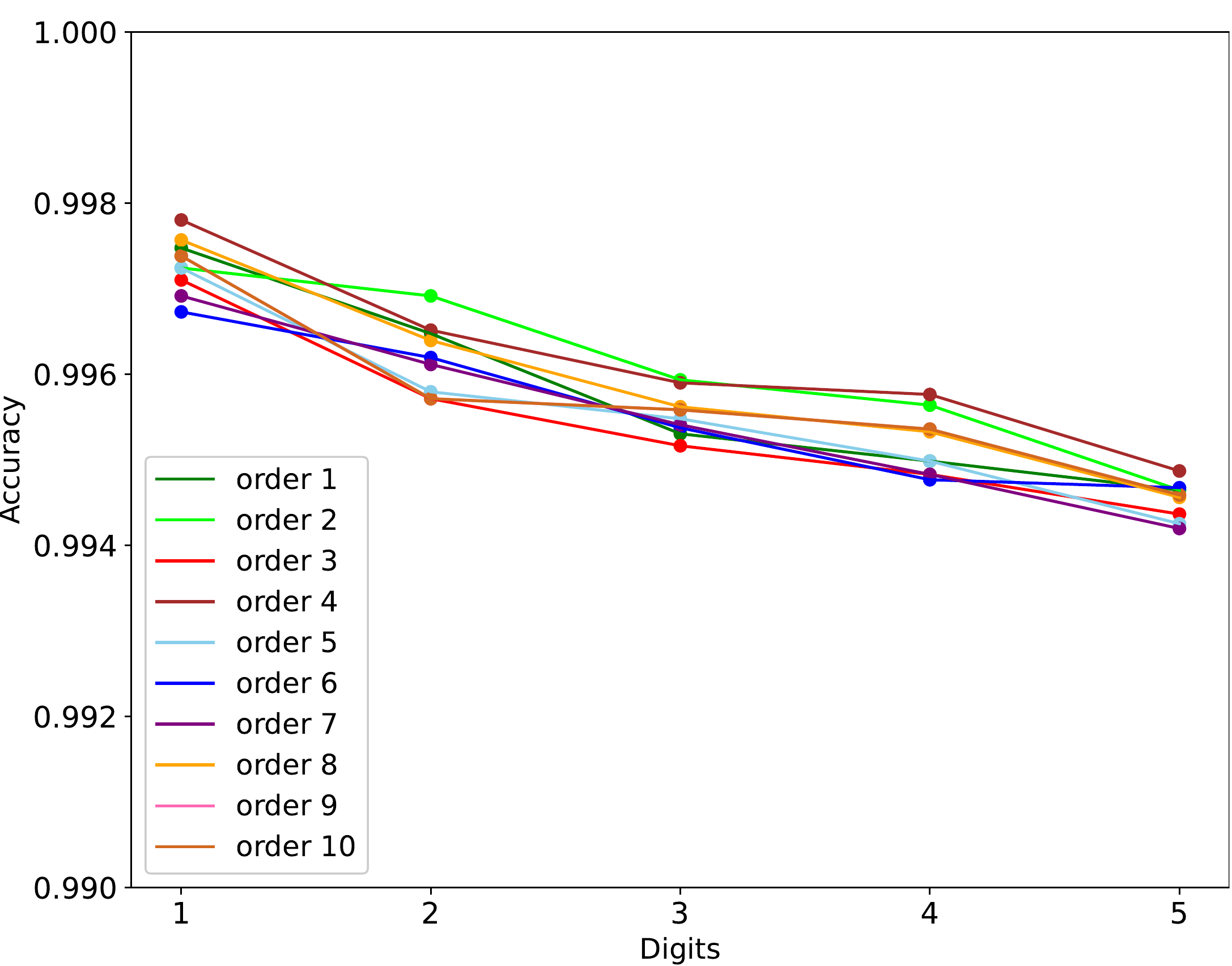} 
	\caption{Bootstrap set $[1,9,3,0,4]$}
	\label{fig:19304}
\end{subfigure} 
\begin{subfigure}{.32\textwidth}
	\centering
	\includegraphics[width=.9\linewidth]{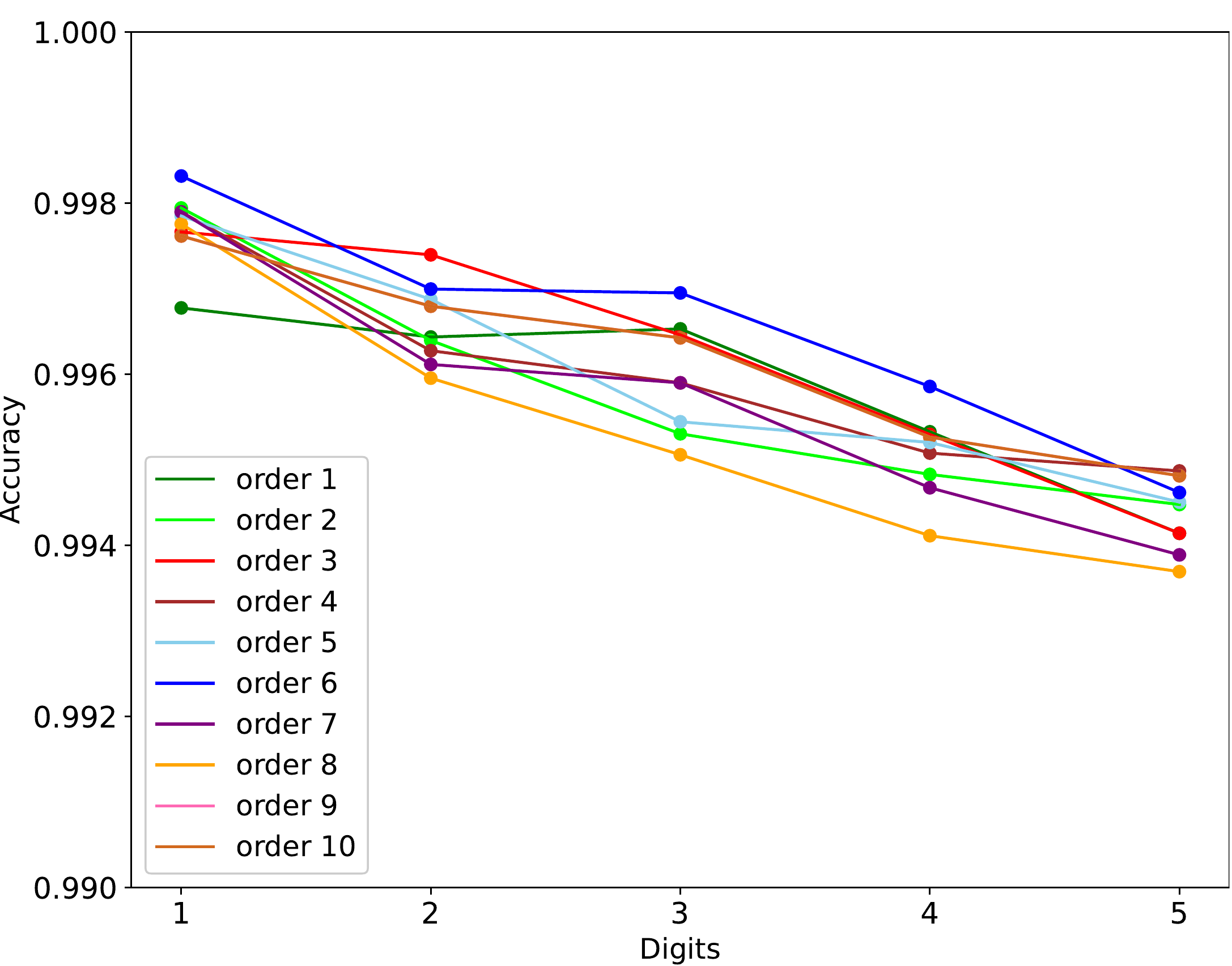} 
	\caption{Bootstrap set $[2,8,7,4,1]$}
	\label{fig:28741}
\end{subfigure}	 \\  \vspace{3mm}
\begin{subfigure}{.32\textwidth}
	\centering
	\includegraphics[width=.9\linewidth]{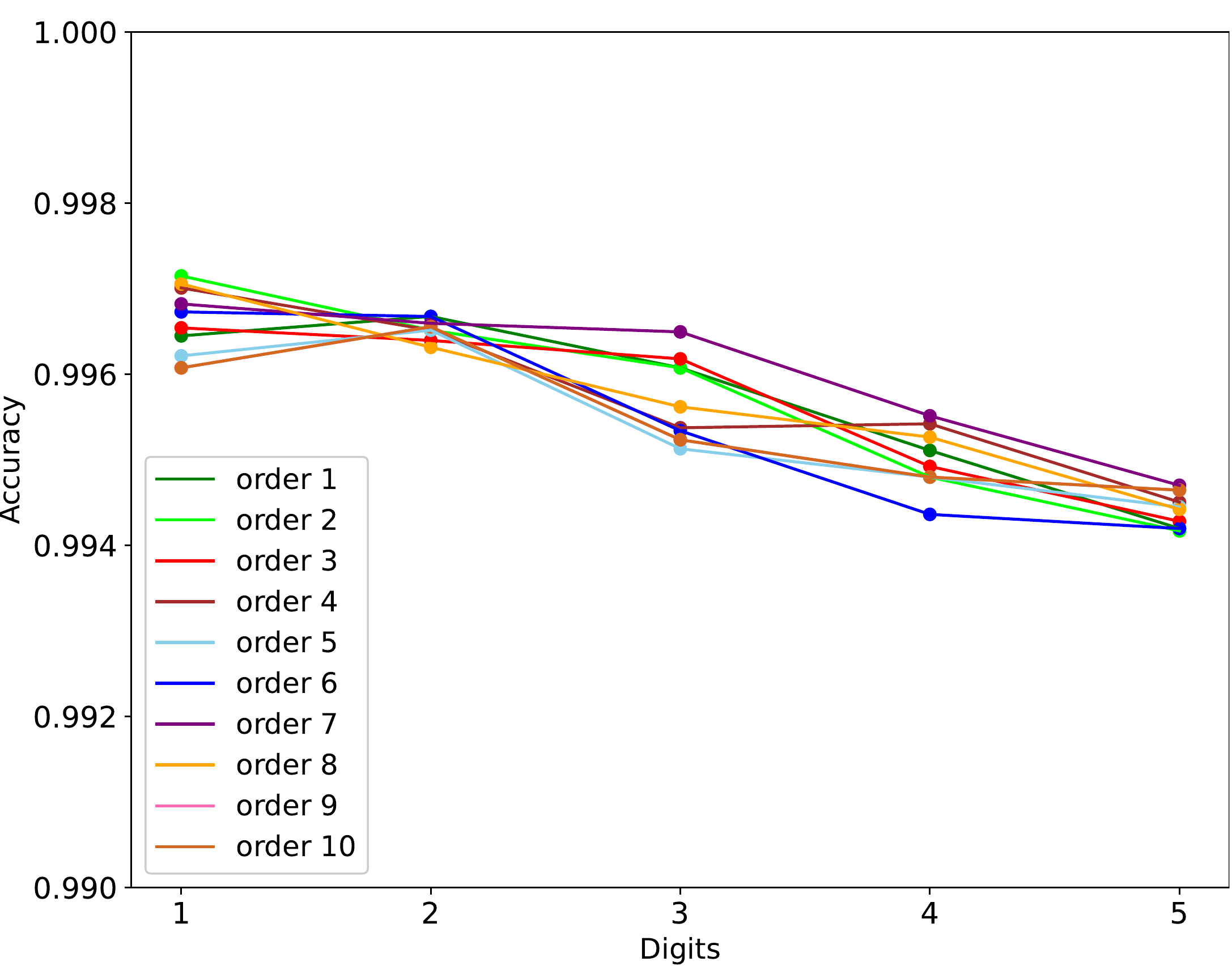} 
	\caption{Bootstrap set $[4,9,6,8,1]$}
	\label{fig:31492}
\end{subfigure} 
\begin{subfigure}{.32\textwidth}
	\centering
	\includegraphics[width=.9\linewidth]{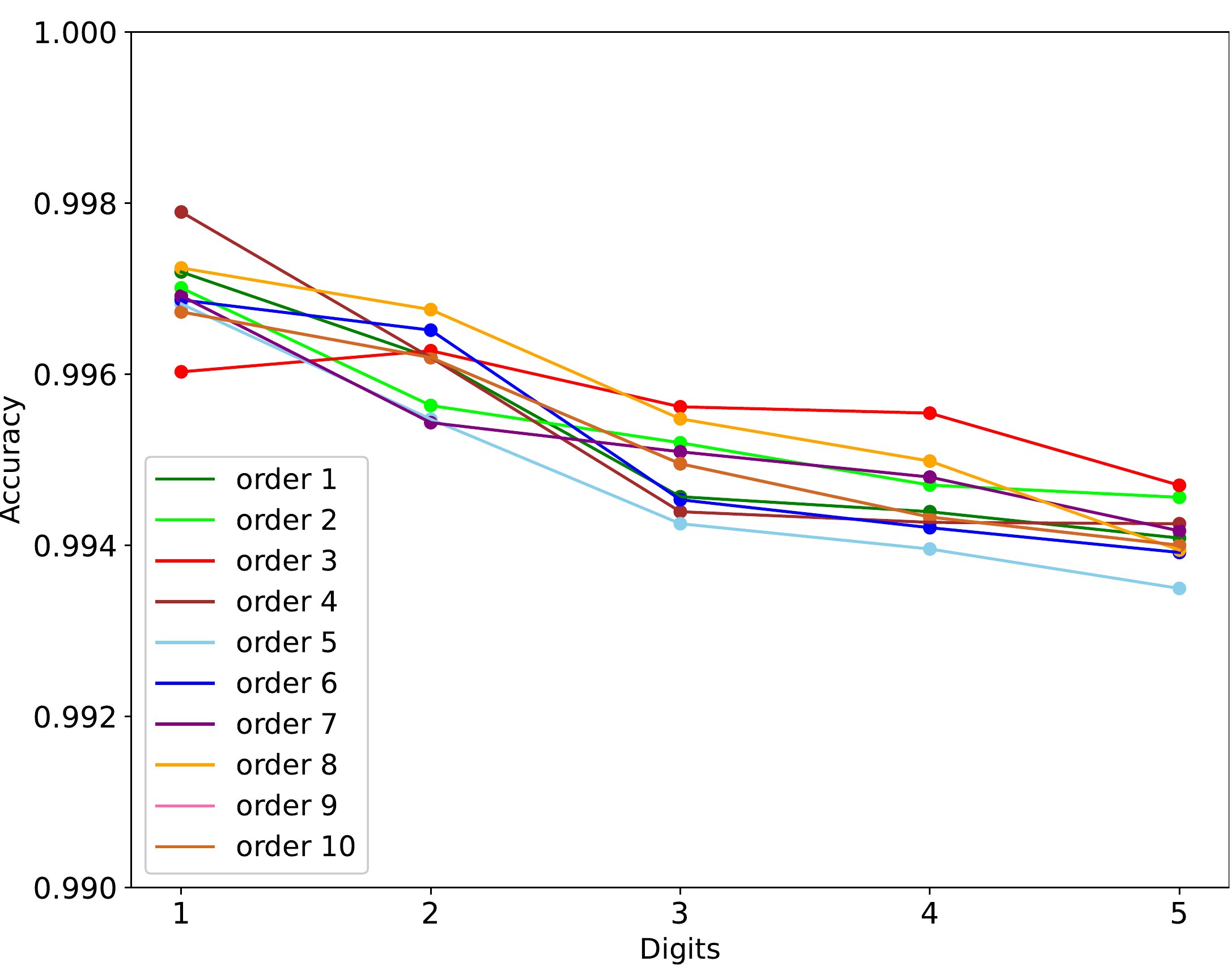} 
	\caption{Bootstrap set $[5,0,4,3,7]$}
	\label{fig:49681}
\end{subfigure}	 
\begin{subfigure}{.32\textwidth}
	\centering
	\includegraphics[width=.9\linewidth]{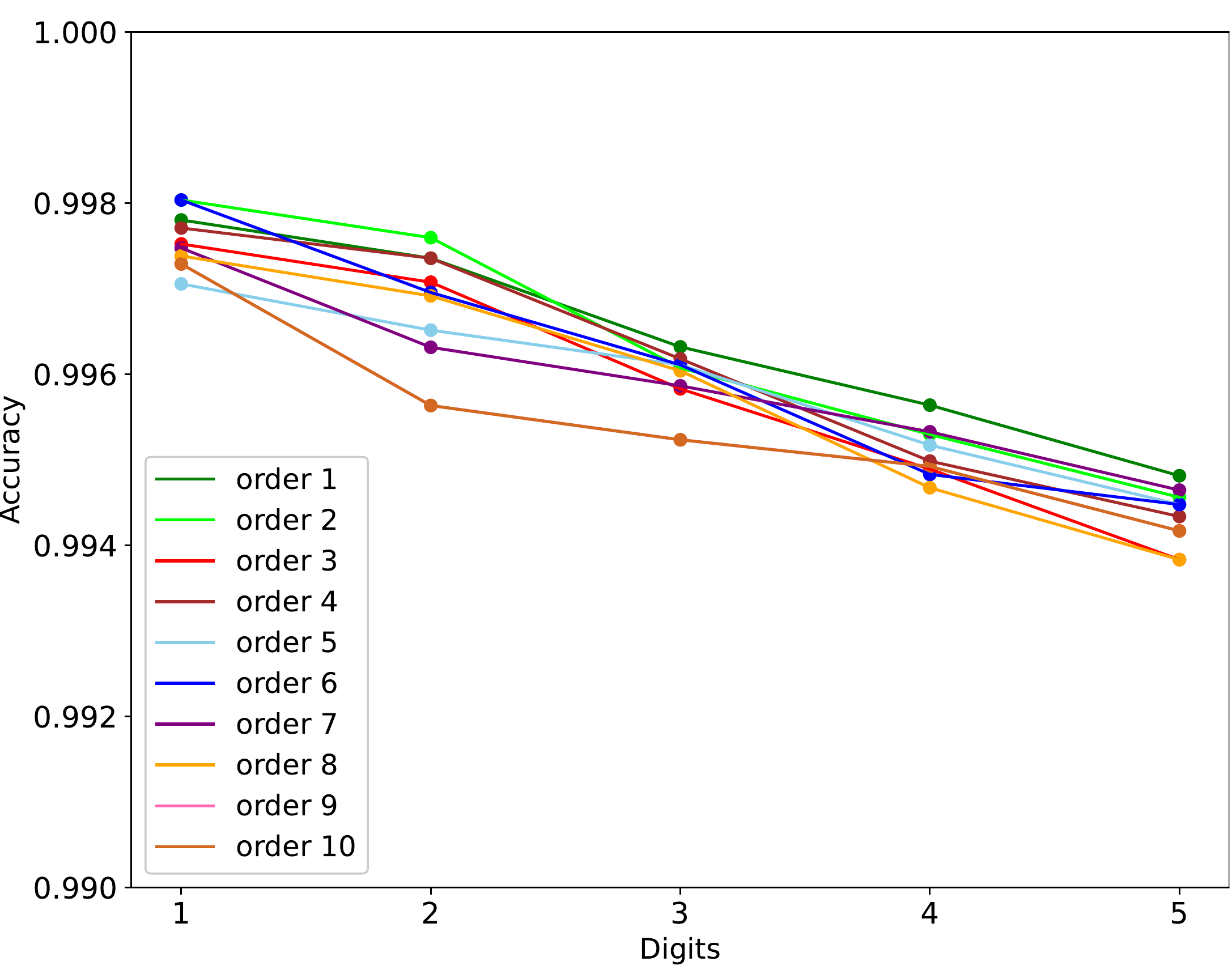} 
	\caption{Bootstrap set $[6,2,0,8,4]$}
	\label{fig:50437}
\end{subfigure} \\  \vspace{3mm}
\begin{subfigure}{.32\textwidth}
	\centering
	\includegraphics[width=.9\linewidth]{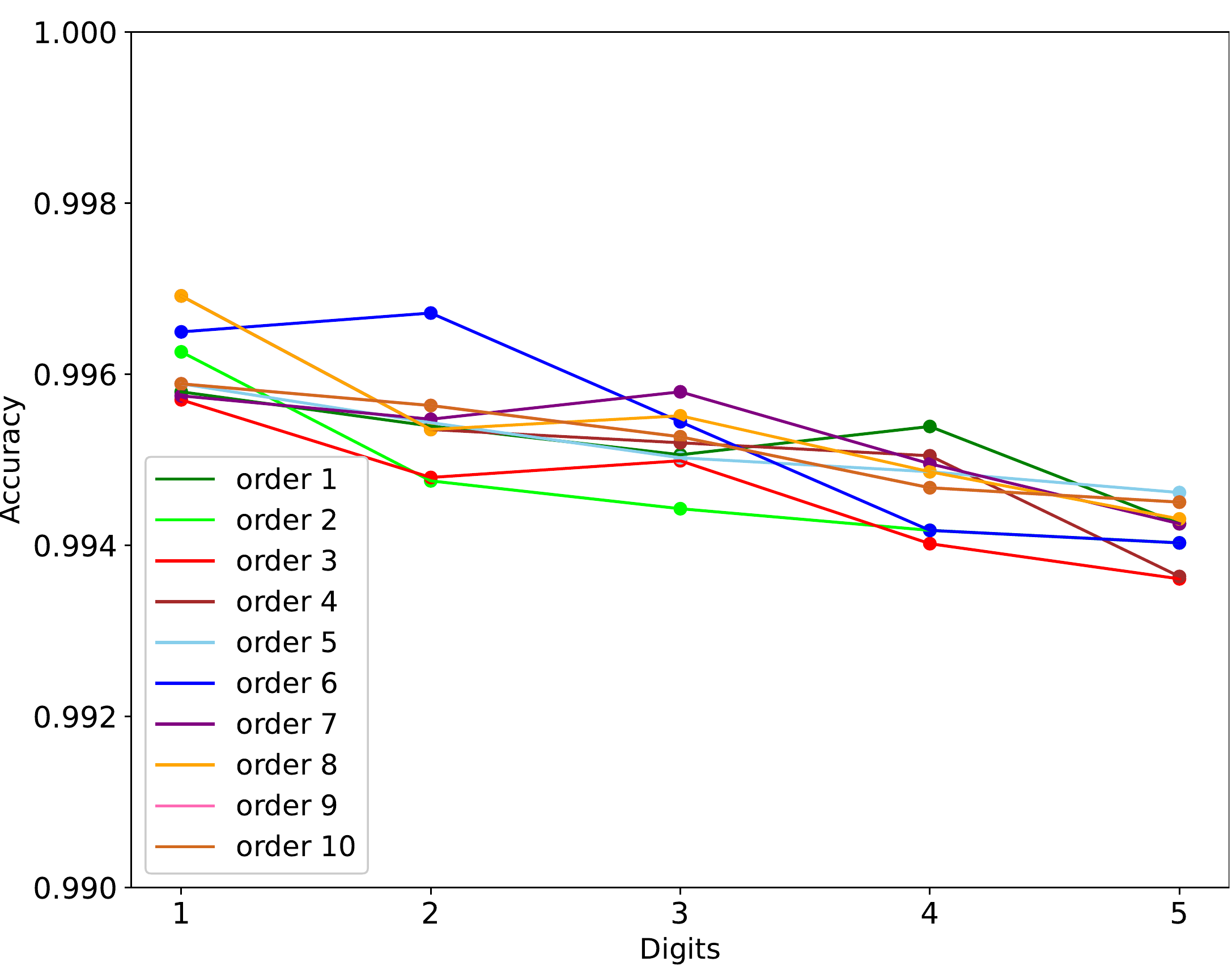} 
	\caption{Bootstrap set $[7,3,1,5,8]$}
	\label{fig:50437}
\end{subfigure} 
\begin{subfigure}{.32\textwidth}
	\centering
	\includegraphics[width=.9\linewidth]{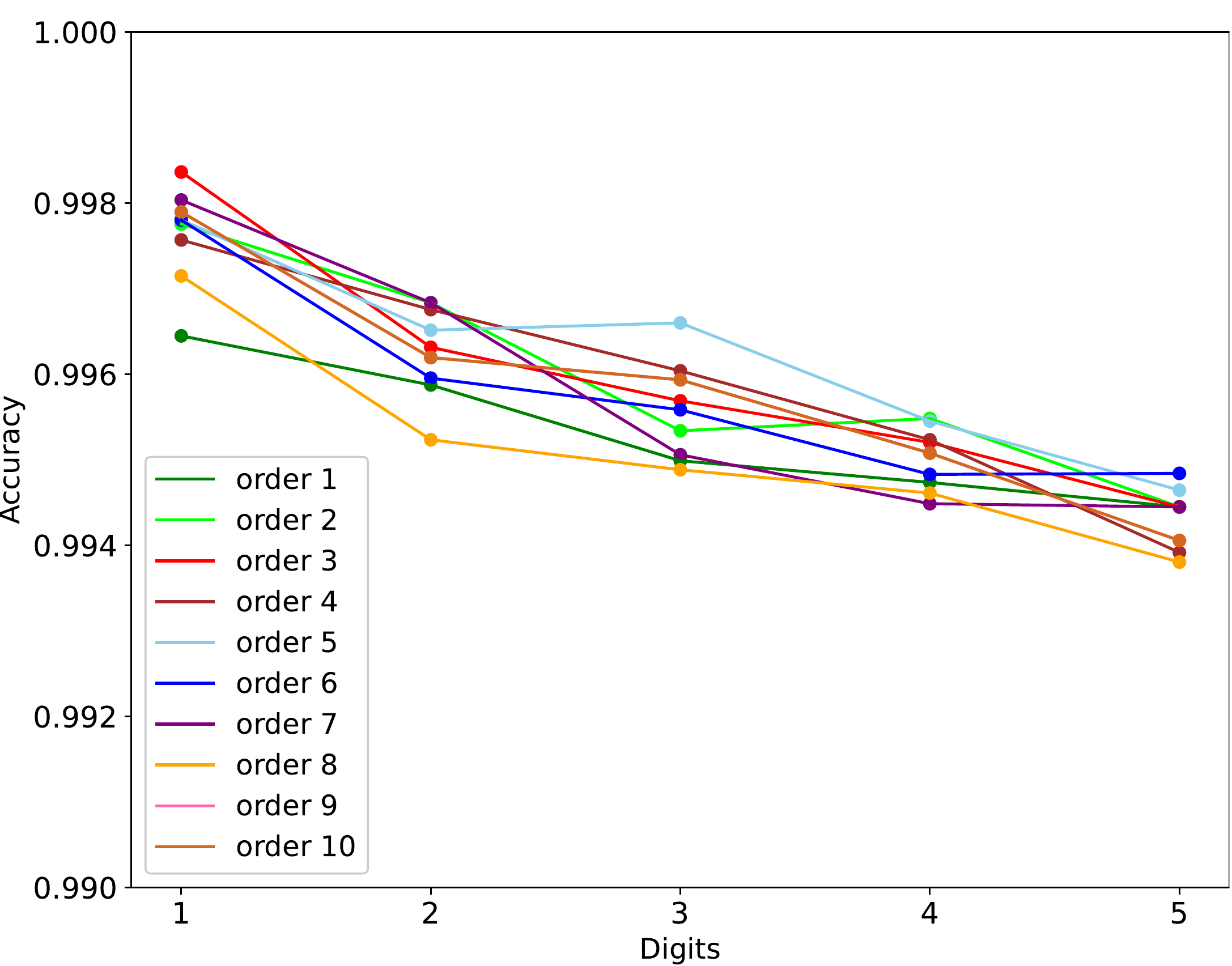} 
	\caption{Bootstrap set $[8,7,2,1,4]$}
	\label{fig:50437}
\end{subfigure} 
\begin{subfigure}{.32\textwidth}
	\centering
	\includegraphics[width=.9\linewidth]{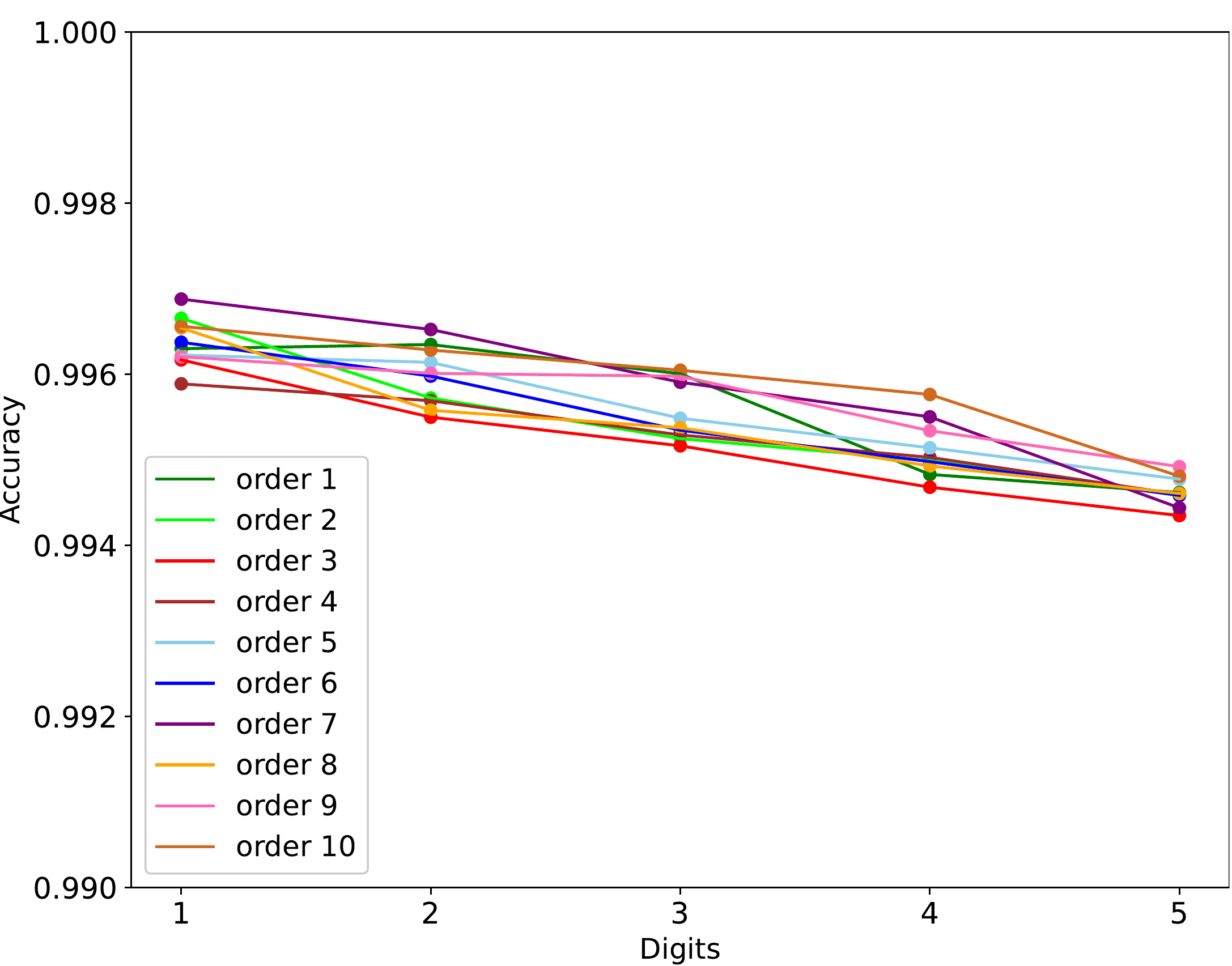}\vspace{-2mm}
	\caption{Bootstrap set $[9,3,4,5,6]$}
	\label{fig:50437}
\end{subfigure} \\ 
\caption{The performance of the accretionary learning model on 10 sequences.\label{fig:Learning_order1}}
\end{figure*} 

Results in Fig. \ref{fig:Learning_order1}   show  the accuracy of the 10 bootstrap sets in Table  \ref{table:1}. Each plot has 10 curves corresponding to the 10 sampled random order of the remaining digits. It is clearly observed that the system performance is largely independent of the learning order. The small accuracy fluctuation due to learning order is remarkably kept within a band of 0.2\%. This confirms the reasoning behind the development of the proposed accretionary learning paradigm and design, as evidenced in this digit recognition experiment.

\subsubsection{Decision Network Replacement}
 This section compares the testing accuracy of the accretionary learning model in different bootstrap sizes using the original and replaced decision network. Training such a replaced decision network is the same as the traditional training method for neural networks. To proceed with that, detectors are trained firstly while the decision network will be kept untrained with detectors. After finishing training ten detectors, unknown weights for the replaced decision network are trained together by using the output of all ten detectors as the input of the decision network. As demonstrated in Fig.\ref{fig:Bootstrap}, it is evident that different bootstrap sizes lead  to  different testing accuracy. In this regard, the bootstrap size still varies from three to ten in order to check the accuracy in the following experiment.

 \begin{figure}[!th]
\centering
\begin{subfigure}{.495\textwidth}
	\centering
	\includegraphics[width=.9\linewidth]{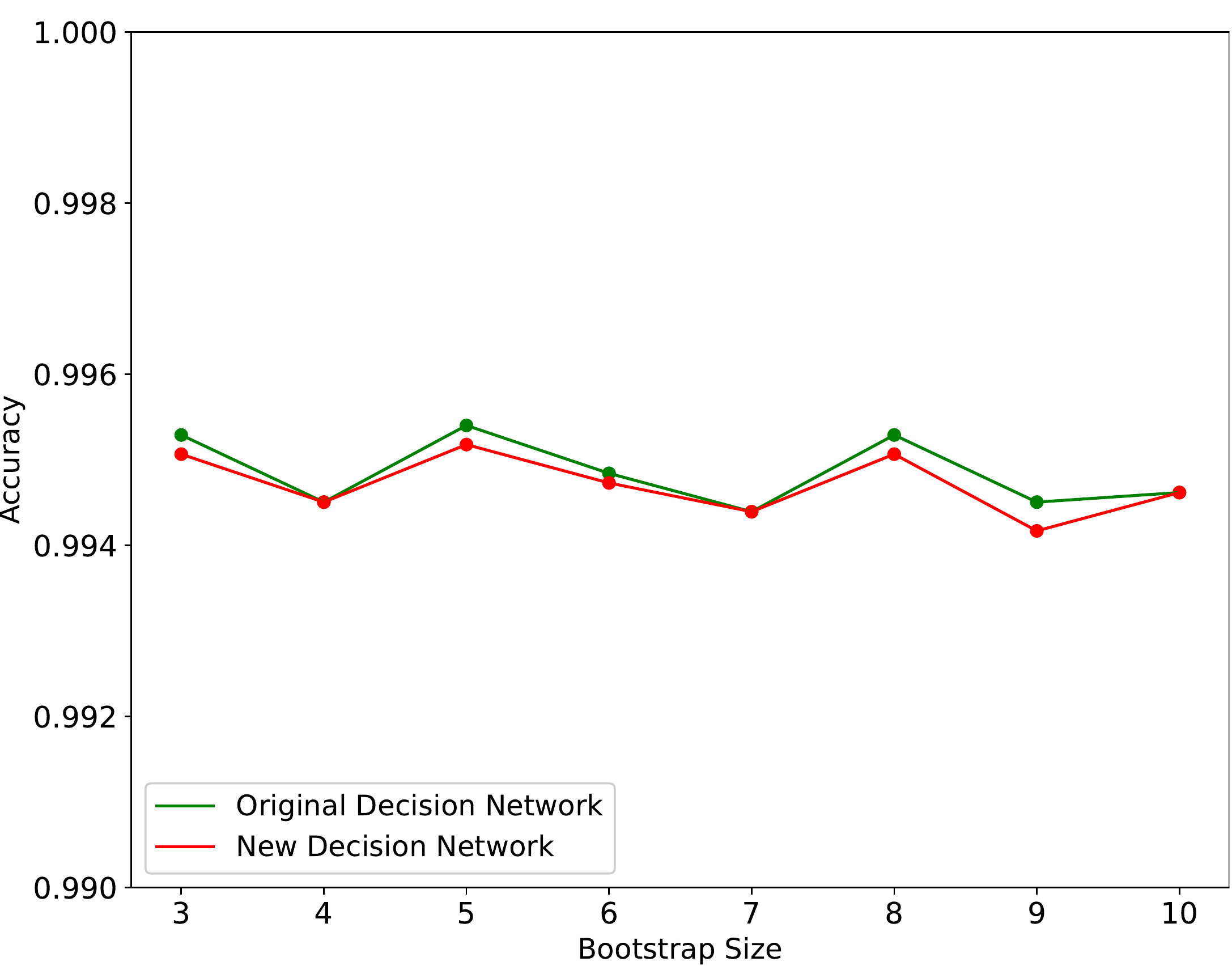} 
	\caption{Original v.s. new replaced decision network}
	\label{fig:original_vs_new_decision}
\end{subfigure}\\	 
\begin{subfigure}{.495\textwidth}
	\centering
	\includegraphics[width=.9\linewidth]{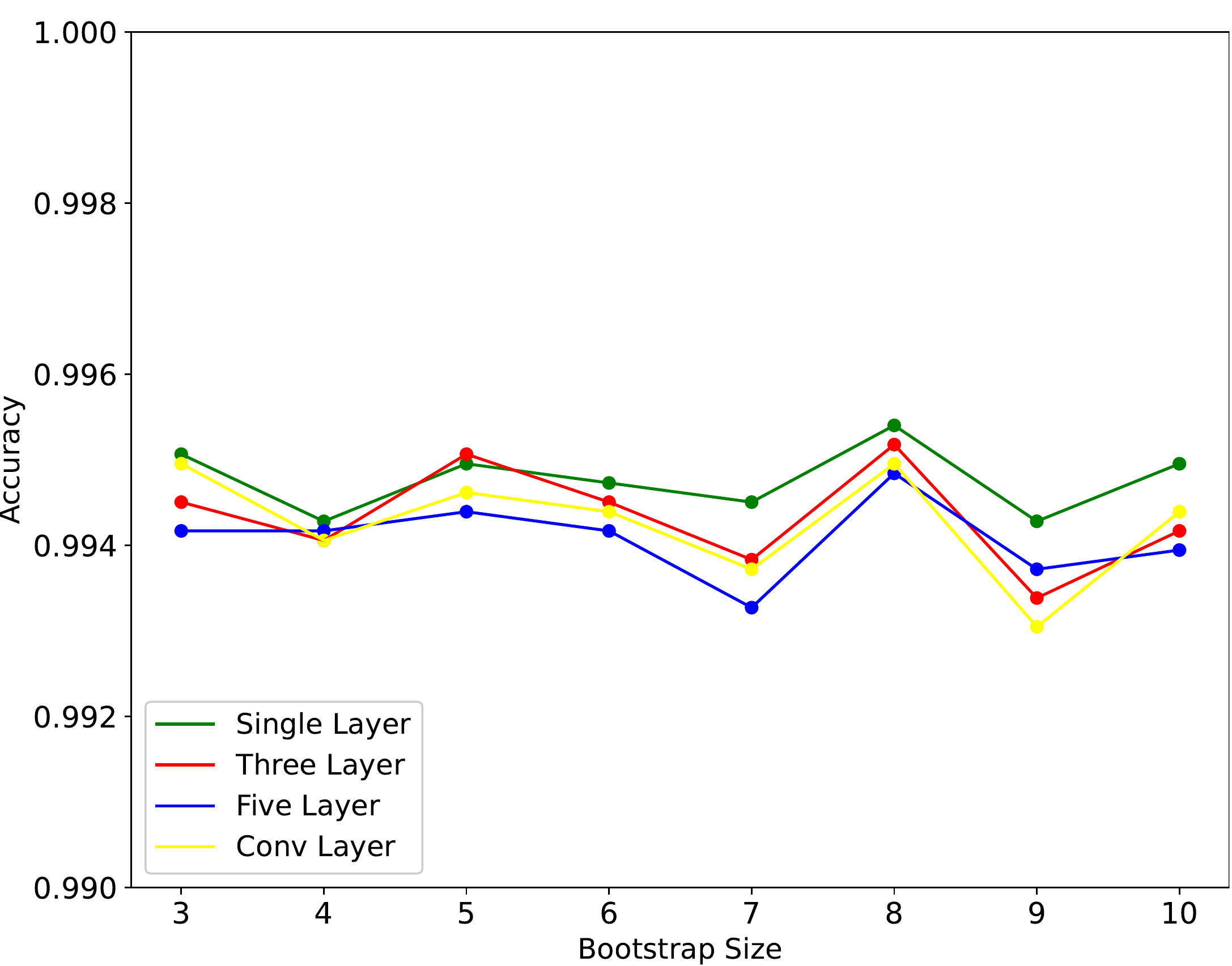} 
	\caption{Four new replaced decision networks}
	\label{fig:multi_model}
\end{subfigure} 
\caption{The testing performance for different structures of the  decision network.}
\end{figure}

 We first consider a simple architecture of the replaced network that is the same as the original decision network, i.e., a single-layer fully-connected neural network. The testing performance for all ten digits with the different bootstrap sizes is shown in Fig.~\ref{fig:original_vs_new_decision}. It can be clearly seen that the testing accuracies of the replaced decision network are bounded (from above) by those of the original decision network. This indicates that such a simple architecture of the replaced network does not improve the accuracy at all.


 Then, three more complex structures of replaced decision networks are taken into account. They are the three-layer fully-connected network, the five-layer fully-connected network, and the two-layer convolutional neural networks. We compare them with the replaced decision network with the single-layer fully-connected network described as above. As shown in Fig.~\ref{fig:multi_model}, the single-layer fully-connected network outperforms the other three replaced decision networks. Overall,  together with the testing results in Fig.~\ref{fig:original_vs_new_decision},  the original decision network behaves the best.
 
 The above results are somewhat surprising and noteworthy. While all networks perform nearly equally well, within a band of 0.001 in accuracy, the single-layer network maintains its superiority. It seems to indicate a particular fitting of the connection weights upon the learning of a new class through the newly computed testing likelihoods, without any perturbation to weights that were learned previously. This needs to be verified through a larger scale study and a task that may demonstrate more substantial performance differences than the current one to allow a more definitive conclusion.

\section{Conclusion}\label{sec:Conclusion}
In this paper, we have applied accretionary learning in deep learning networks. The corresponding model structure is able to learn new data classes continually without re-designing and retraining processes. During accretionary learning in new data classes in MNIST dataset, our accretionary model shows stable performance that it will not forget the previous knowledge while having great performance on recognizing the new data classes. Our future work will concentrate on looking into and improving the performance of each module of our model in order to make the overall system work more efficiently.


%

\bibliographystyle{IEEEtran}
\bibliography{references}



%








\end{document}